\documentclass[letterpaper, 10 pt, conference]{ieeeconf}

\usepackage{amsmath}
\usepackage{amssymb}
\usepackage{bm}
\usepackage{graphicx}
\usepackage{enumerate}
\DeclareMathOperator*{\E}{\mathbb{E}}
\usepackage{todonotes}
\usepackage{algorithm}
\usepackage{algpseudocode}

\usepackage{subcaption}
\usepackage{adjustbox}
\usepackage{url}
\usepackage{caption}
\usepackage{hyperref}
\usepackage[capitalise]{cleveref}
\newcommand{\bb}{\mathbf}  

\IEEEoverridecommandlockouts
\title{\LARGE \bf
Teaching Robots Where To Go And How To Act With Human Sketches\\ via Spatial Diagrammatic Instructions
}

\author{
    Qilin Sun$^{1,*}$ \and Weiming Zhi$^{1,*}$ \and Tianyi Zhang $^{1}$ \and Matthew Johnson-Roberson $^{1}$%
    \thanks{$^{*}$ Equal Contribution}%
    \thanks{$^{1}$ Robotics Institute, Carnegie Mellon University, PA, USA.}
    \thanks{Code is available at \url{https://github.com/PtSamuel/diagrammatic_instructions}}
}

\begin{document}

\maketitle
\thispagestyle{empty}
\pagestyle{empty}

\begin{abstract}
This paper introduces \emph{Spatial Diagrammatic Instructions} (SDIs), an approach for human operators to specify objectives and constraints that are related to spatial regions in the working environment. Human operators are enabled to sketch out regions directly on camera images that correspond to the objectives and constraints. These sketches are projected to 3D spatial coordinates, and continuous \emph{Spatial Instruction Maps} (SIMs) are learned upon them. These maps can then be integrated into optimization problems for tasks of robots. In particular, we demonstrate how Spatial Diagrammatic Instructions can be applied to solve the Base Placement Problem of mobile manipulators, which concerns the best place to put the manipulator to facilitate a certain task. Human operators can specify, via sketch, spatial regions of interest for a manipulation task and permissible regions for the mobile manipulator to be at. Then, an optimization problem that maximizes the manipulator's reachability, or coverage, over the designated regions of interest while remaining in the permissible regions is solved. We provide extensive empirical evaluations, and show that our formulation of Spatial Instruction Maps provides accurate representations of user-specified diagrammatic instructions. Furthermore, we demonstrate that our diagrammatic approach to the Mobile Base Placement Problem enables higher quality solutions and faster runtime.     
\end{abstract}

\section{Introduction}
We envision a future where human operators work jointly with robots. To achieve this, human operators need to be able to readily deliver instructions to the robot. These inputs can take many modalities, including language instructions or kinesthetic demonstrations. Each of these modalities has its benefits and limitations. Language is a natural and humanesque way to provide commands to a robot, but would require brittle pipelines to combine the language input with semantic information from its perception system. Kinesthetic demonstrations can be obtained when the user is co-located with a fixed-based manipulator. However, the physical handling of a mobile manipulator can be challenging, if not impossible.   

\begin{figure}[t]
    \centering
    \includegraphics[width=0.45\textwidth]{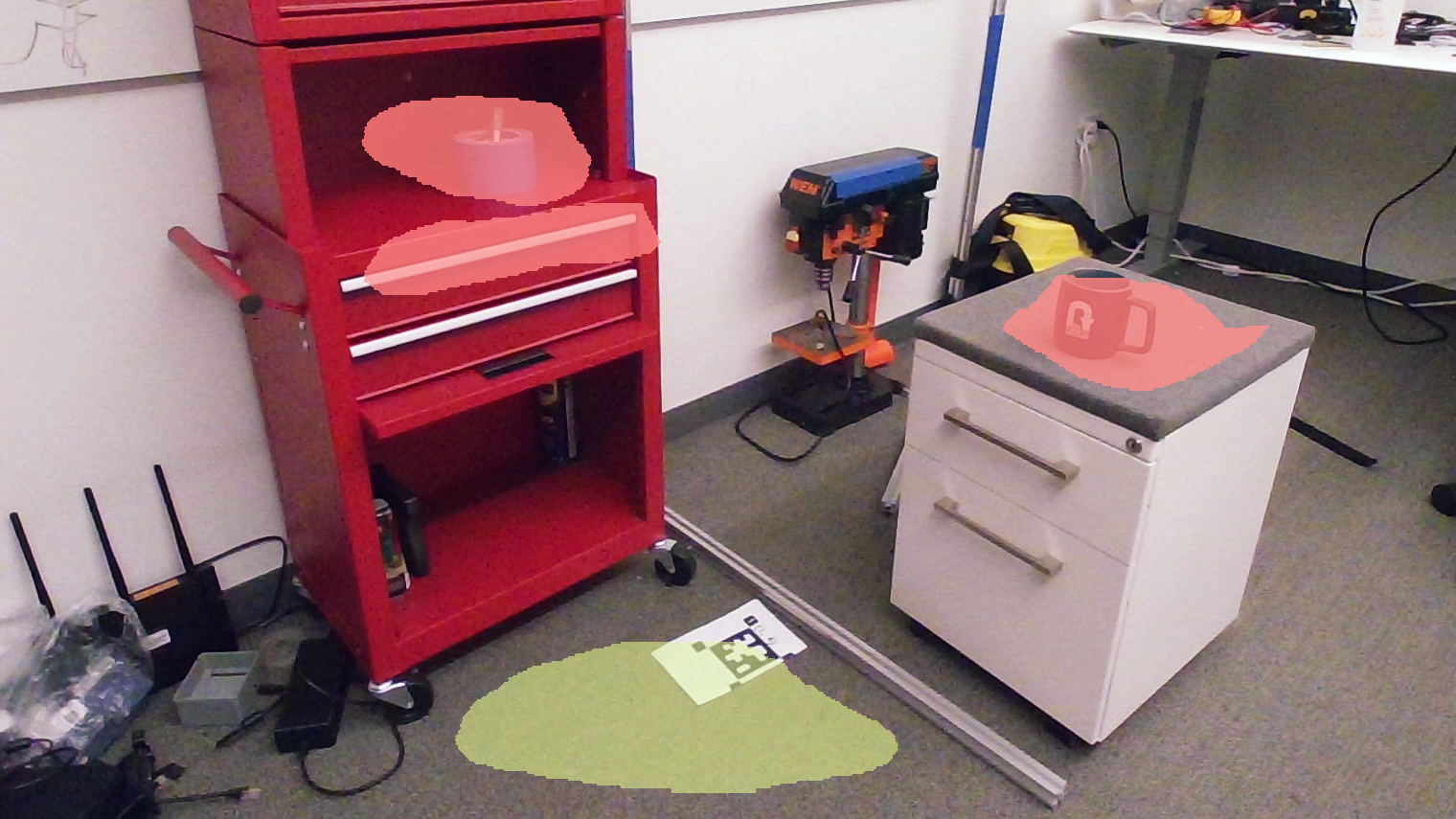}%
    \vspace{0.1em}
    \includegraphics[width=0.45\textwidth]{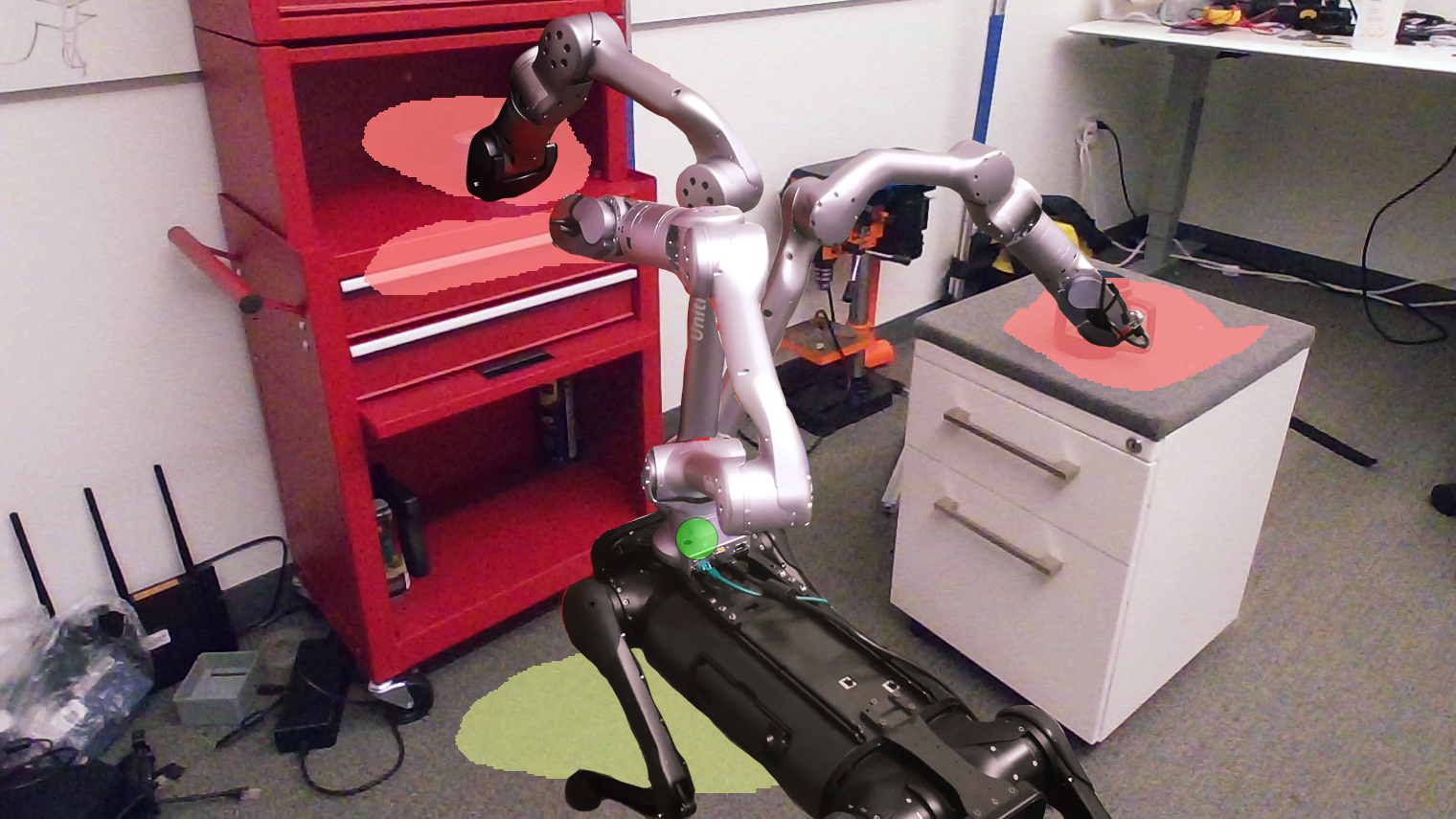}
    \caption{Spatial Diagrammatic Instructions enable users to sketch on camera images of the environment. Top: Red indicates regions where the user sketches as regions of interest, while green indicates regions the user sketches as permissible regions. Bottom: Continuous spatial representations of these sketched regions can then be incorporated as objectives and constraints within optimization problems to find optimal positions for mobile manipulator base placement.}
    \label{fig:instructions}
\end{figure}

In this paper, we develop a \emph{Spatial Diagrammatic Instructions} (SDIs) framework, where the user directly sketches over images from camera views of the environment to identify regions that are relevant for an instruction. We highlight that humans have the exceptional ability to make inferences from diagrammatic sketches over images and relate instructions with spatial regions in the physical environment. We seek to endow collaborative robots with this same capability. We propose \emph{Spatial Instruction Maps} (SIMs) which take user-specified sketches to learn differentiable and probabilistic models that identify the corresponding 3D spatial regions. 

In particular, we leverage diagrammatic instructions to enable robot-human collaboration to solve for the optimal placement of the mobile base. That is, where should a mobile manipulator navigate to, such that it can effortlessly reach objects in the regions of interest? We demonstrate how our proposed spatial representations are amenable to integration into optimization problems, with the differentiability of the representations facilitating efficient optimization.    

Concretely, the technical contributions of this paper include:
\begin{enumerate}
    \item The \emph{Spatial Diagrammatic Instructions} framework for human operators to specify spatial regions for optimization problems;
    \item \emph{Spatial Instruction Maps}, a \emph{probabilistic} and \emph{continuous} representation of user-specified regions as energy-based models;
    \item An inverse kinematic-free method that leverages user diagrammatic specifications to solve base placement for a mobile manipulator. 
\end{enumerate}

The remainder of the paper is as follows: we shall first give a bird's-eye view summary of the past work that is relevant to this paper in \cref{sec:related}. Then, we introduce the concept of \emph{Spatial Diagrammatic Instructions}, which is central to our contributions, along with the problem formulation in \cref{sec:formulation}. Next, we describe the contributed methodology of building \emph{Spatial Instruction Maps}, along with the integration of the maps into an optimization problem to solve the mobile base placement problem in \cref{sec:opt}. We subsequently report empirical evaluation of our contributed methodology in \cref{sec: experiments}  and conclude in \cref{sec:conc}. 

\section{Related work}\label{sec:related}

\textbf{Robot Learning from Human Input:}
Humans can provide robots a wealth of information around how to act. One of the most well-studied problems where robots learn from human inputs is imitation learning \cite{ravichandar2020recent,promp,learning_mps}. In these setups, a human expert will provide demonstrations of the desired movement and the robot seeks to mimic the motion patterns. Frameworks around learning and then importantly generalizing the motion have also been proposed \cite{Diff_templates,clamp}. Other extensions to imitation learning that consider human preference, where desirable behaviors are strengthened while undesirable ones are diminished, have also been explored \cite{NIPS2013_e034fb6b}. Specifying robot motions via natural language has also been an area of research \cite{Tellex2020RobotsTU}. Diagrammatic Learning \cite{diagrammatic} has been proposed as a medium to teach robots motion by sketching the desired motion. Similarly, this work seeks to enable users to provide input via sketches, but of desired spatial regions rather than motion trajectories. 

\textbf{Building Spatial Representations:}
Representing the environment where a robot operates is central to robotics. Early approaches in this area looked at grid maps \cite{OccupancyGridMaps, THRUN199821}. Various approaches have also been explored to build continuous environment representations for robotics. These include models of occupancy \cite{HM, HilbertMaps}, motion \cite{sptemp}, distance \cite{Park_2019_CVPR}, and photo-realistic models \cite{mildenhall2020nerf, mueller2022instant, zhang2024darkgs}. In this work, we seek to construct continuous representations that are amenable to optimization, but the properties we wish to embed into the representation are spatial instructions that can be used for planning \cite{lavalle:2006,PDMP} and optimization problems \cite{chomp, GeoFab_gloabL_opt, Fast_diff_int}.

\textbf{Base Placement for Mobile Manipulators:}
Positioning mobile manipulators has received notable interest in the context of industrial manufacturing. A painting robot, for example, usually paints a large work-piece portion by portion. A trajectory for the end-effector pose is given for each portion, and the base position needs to be optimized accordingly \cite{YU201856}. Under scenarios like this, multiple performance indices that serve as the optimization object have been proposed. One intuitive way is to maximize the proportion of target points that are achievable \cite{yang2009placement}. Besides, \cite{YU201856} proposed optimizing for the manipulability and dexterity of joints, \cite{FAN2021102138} incorporated the manipulator stiffness into the objective, and \cite{GADALETA2017102} aimed at minimizing the energy consumption for completing a certain task. Within the robotics community, two methodologies have been explored: using a reachability map vs. not using one. For the former, \cite{6630839} prepared a reachability map offline, by discretizing the space of end-effector poses. \cite{8329892} extended this methods by applying a clustering algorithm on the potential base poses to find the best ones. For the other approach that does not use a reachability map, \cite{4543365} proposed a co-evolutionary method to optimize the base positions for a pick-and-place task, combining grasp quality, manipulability, and collision avoidance into the performance metric. Unlike our proposed approach, these methods require an \emph{a priori} representation of the environment layout and do not allow the user to specify desired objectives and constraints onto images.  

\section{Problem Formulation}\label{sec:formulation}
In this section, we begin by introducing the concept of \emph{Spatial Diagrammatic Instructions} (SDIs). Then, we define the \emph{Mobile Base Placement Problem} (MBPP) and outline how the SDIs enable humans to collaborate with the robot to solve the MBPP.

\subsection{Spatial Diagrammatic Instructions}

We are assumed to have access to an RGB-D camera with known intrinsics and pose. We begin by providing the human operator with an RGB image of the environment, then prompt the human operator to provide spatial instructions by outlining a region on the image. Given the camera parameters, we project the instruction-specified regions into the environment, i.e. $\mathcal{X}_{1},\mathcal{X}_{2},\ldots, \mathcal{X}_{m}\subseteq \mathbb{R}^{3}$, where $m$ is the number of instructions. Then, we can leverage the specified regions within optimization problems in robotics. These take the general form:
\begin{align}
&&\mathrm{Optimize}
\quad F(x| \mathcal{X}_{1},\mathcal{X}_{2},\ldots, \mathcal{X}_{m})\\
\text{s.t.} && \quad G(x|\mathcal{X}_{1},\mathcal{X}_{2},\ldots, \mathcal{X}_{m}) = 0,
\end{align}
where $F$ is an objective function and $G$ denotes a set of constraint functions.

\subsection{Mobile Base Placement Problem}
A particular optimization problem that can benefit from diagrammatic instructions is the \emph{Mobile Base Placement Problem} (MBPP). Consider the placement of a mobile base with a mounted manipulator, such that it has maximum coverage over a region of interest (ROI). This ROI should include regions where manipulation tasks are likely to begin or end. For example, in the task of moving objects from one table to another, the ROI would be the 2 tabletops. Additionally, the mobile base can also be constrained to be placed within a designated region. 

The motivation for this problem is described as follows. When a robot manipulator mounted on a mobile platform performs a task like picking and placing an object, both the robot manipulator and the platform may move. Allowing motion of both, however, increases the difficulty in planning and potentially leads to greater actuation error. Where possible, it is therefore desirable to keep the mobile platform fixed in place, and let the manipulator complete the whole task on its own. In this work, we attempt to find the base placement that best facilitates downstream tasks, especially those that require the manipulator to move long ranges. We now introduce the notations used throughout the paper. Let $\mathcal{X}_{ROI}$ denote the ROI that the mobile manipulator seeks to cover, and let the mobile base configuration be $\mathcal{C}$. In this paper, we consider $\mathcal{C}$ to have 3D coordinates $\bb{x} \in \mathbb{R}^3$ and yaw angle $\omega$ as states, i.e. $\mathcal{C} = (\bb{x}, \omega)$, but $\mathcal{C}$ can also be defined more generally. Let $\mathcal{B}$ denote the set of all allowable base configurations $\mathcal{C}$. We also need the forward kinematics of the manipulator $\bb{x}_{e} = f_{e}(\bb{q} \mid \mathcal{C})$, where $\bb{x}_{e}\in\bb{SE}(3)$ is the end-effector pose, which is a function of both $\bb{q}$, the manipulator's joint configurations, and the base configuration $\mathcal{C}$. We view the optimal base placement as one that maximizes the coverage of some ROI, and we quantify the coverage by the probability $\bb{P}(f_{e}(\bb{q} \mid \mathcal{C}) \in \mathcal{X}_{ROI})$. Therefore, we define the \emph{Mobile Base Placement Problem} (MBPP) as:
\begin{align}
\textbf{[MBPP]:} \qquad & \arg \max \limits_{\mathcal{C}} \bb{P}(f_{e}(\bb{q} \mid \mathcal{C}) \in \mathcal{X}_{ROI}), \label{eq:obj_prob} \\
\text{s.t.} \qquad & \mathcal{C}\in \mathcal{B}.
\end{align}
Note that we leave $\bb{q}$ as a free variable, which will be discussed in \cref{subsec:solving}.

\section{Optimization on Spatial Instruction Maps} \label{sec:opt}
After the human operator is prompted to sketch Spatial Diagrammatic Instructions over camera images, the instructions are projected into 3D space (outlined in \cref{subsec:region}). Then, we construct concise and continuous Spatial Instruction Maps using \emph{energy-based models} (\cref{subsec:ebm}). These models are subsequently integrated into the mobile base placement problem to be solved. Solving the optimization is detailed in \cref{subsec:solving} and the enforcement of constraints is elaborated in \cref{subsec:constraints}.  

\subsection{Projecting Spatial Diagrammatic Instructions into 3D}
\label{subsec:region}
The human operator is prompted to sketch instructions directly onto images of the environment captured by an RGB-D camera and enclose regions on the image relevant to an instruction. Though these diagrams are 2D, they relate to relevant spaces within the 3D environment, and form the basis of optimization objectives (e.g. coverage of the selected space) and constraints (e.g. regions to stay in or stay out of). 

After the operator has sketched the diagrammatic instructions, we then obtain all the pixels that are enclosed in the sketch and find their projection into the 3D environment. Here, we denote each pixel projected into 3D as $\bb{x}_{s} \in  \mathcal{S} \subseteq \mathbb{R}^{3}$, where $\mathcal{S}$ denotes the set of 3D points corresponding to the region enclosed by the sketch. We then have, for each enclosed pixel $(u,v)$, the transformation:
\begin{equation*}
    \bb{x}_{s} = R_C \begin{bmatrix}
        \frac{u - c_x}{f_x} z \\
        \frac{v - c_y}{f_y} z \\
        z
    \end{bmatrix} + t_C,
\end{equation*}
where $z$ is the measured depth, or the distance from the pixel to the camera's sensor plane, $R_C, t_C$ are respectively the camera's rotation and translation in the world frame, and $f_x, f_y, c_x, c_y$ are the focal lengths and center points along $x$ and $y$ axes of the camera. These parameters can be obtained from the camera's manufacturer. After this is done, all of the 3D points obtained, $\bb{x}_{s} \in \mathcal{S}$, are scattered in the regions outlined by the human operator. Next, we shall explore the construction of models defined over these regions. 

\subsection{Spatial Representations via Energy-Based Models}
\label{subsec:ebm}

Diagrammatic instructions are sketched onto images with discrete pixels which are subsequently projected as a discrete point set $\mathcal{S}$ into 3D space, but optimization objectives and constraints defined over the corresponding specified regions require a continuous domain. We therefore need to develop continuous and probabilistic representations of the specified regions. Here, we model the probability of any coordinate $\bb{x}$ being in the space specified by the user in the sketches, $\bb{P}(\bb{x} \in \mathcal{S} \mid \bb{x})$, as a continuous function over $\bb{x}\in\mathbb{R}^{3}$, which we call the \emph{Spatial Instruction Map} (SIM).

We choose to use \emph{energy-based models}, which are highly flexible and can model complex distributions efficiently. Energy-based models (EBMs) are probabilistic models with densities over data $\bb{x}$ given as:
\begin{align}
p_{\theta}(\bb{x})=\frac{\exp(E_{\theta}(\bb{x}))}{Z_{\theta}}, && Z_{\theta}=\int\exp(E_{\theta}(\bb{x}))\mathrm{d}\bb{x},
\end{align}
where $E_{\theta}(\bb{x})\in\mathbb{R}$ is known as the \emph{energy function}, which is here represented as a neural network with parameters $\theta$, and $Z_{\theta}$ is the normalizing constant (or \emph{partition function}).

Noise-contrastive approaches provide a way to estimate parameters of the EBM, $\theta$, without direct access to $Z_{\theta}$ by learning a classifier to identify data from generated negative samples. In particular, \emph{Noise Contrastive Estimation (NCE)} \cite{Gutmann2010NoisecontrastiveEA} is an approach that assumes that there are negative examples $\bb{\tilde{x}}$ drawn from a noise distribution $q(\bb{\tilde{x}})$. The energy estimation is then cast as a binary classification problem, and a logistic regressor is applied to minimize the binary cross-entropy loss:
\begin{align}
    \mathcal{L}_{\mathrm{NCE}}(\theta)&=\sum^{N}_{i=1}[\log\sigma(E_{\theta}(\bb{x}_{i}))+\log(1-\sigma(E_{\theta}(\bb{\tilde{x}}_{i})))], \label{eqn:NCE_loss}
\end{align}
where $\sigma(\cdot)$ is the sigmoid function. The loss function can be efficiently optimized via gradient descent methods. Consequently, we can train $E_\theta(\cdot)$ as a classifier that identifies points in $\mathcal{S}$ from generated negative samples, then the trained model will represent $\mathcal{S}$ as a continuous, differentiable distribution. And importantly, because $E_\theta(\cdot)$ is trained as a classifier, we have
\begin{equation}
    \bb{P}(\bb{x} \in \mathcal{S} \mid \bb{x}) = \sigma(E_{\theta}(\bb{x}_{i})). \label{eq:energy_to_prob}
\end{equation}

$E_\theta(\cdot)$ is trained as follows: $E_{\theta}(\cdot)$ is modeled by a 2-layer fully connected neural network with width 256. Positive samples are uniformly drawn from $\mathcal{S}$, and negative samples by uniformly sampled from a larger space surrounding $\mathcal{S}$. All training is conducted with an AdamW optimizer \cite{loshchilov2019decoupled} with learning rate $10^{-3}$ and weight decay $10^{-4}$.

\subsection{Solving the MBPP} \label{subsec:solving}
After preparing the SIMs based on EBM, we gain a continuous representation of the ROI and can now proceed to solving the MBPP. A commonly used approach to find the joint configurations and base configuration to reach a target pose is by an inverse kinematics (IK) solver. In the current setting, however, IK-based methods have a fundamental limitation. The unconstrained end-effector orientation leaves significantly more freedom for the solver, which is compounded by the kinematic redundancies introduced by the mobile base. Consequently, to achieve an end-effector position, there could be a wide ranges of allowable base placements, and selecting the base placement that maximizes the coverage of the ROI could be difficult, if not intractable. 

To address this problem, we model the reachable positions by the end-effector as a probability distribution over all the allowable joint configurations, $\mathcal{Q}$, and let $p_{\bb{q}}(\cdot)$ be the joint p.d.f. of $\bb{q}$. Recall from \cref{eq:obj_prob} that given a base configuration $\mathcal{C}$, the probability of end-effector lying in the ROI is:
\begin{align}
    \bb{P}(f_{e}(\bb{q} \mid \mathcal{C}) \in \mathcal{X}_{ROI}) = \E\limits_{\bb{q} \in \mathcal{Q}}(\bb{P}(f_{e}(\bb{q} \mid \mathcal{C}) \in \mathcal{X}_{ROI})),
\end{align}
which is the expectation over all $\bb{q} \in \bb{\mathcal{Q}}$. By \cref{eq:energy_to_prob}, we have 
\begin{equation}
    \bb{P}(f_{e}(\bb{q} \mid \mathcal{C}) \in \mathcal{X}_{ROI}) = \E\limits_{\bb{q} \in \mathcal{Q}} (\sigma(E_\theta(f_{e}(\bb{q} \mid \mathcal{C})))).
\end{equation}
Optimization on $E_\theta(\bb{x})$ is more robust than on $\sigma(E_\theta(\bb{x}))$, so we instead estimate the expected energy,
\begin{align*}
    \mathcal{E}(\mathcal{C}) = \E\limits_{\bb{q} \in \mathcal{Q}} (E_{\theta}(f_e(\bb{q} \mid \mathcal{C}))).
\end{align*}

As $p_{\bb{q}}(\cdot)$ is unknown, we assume the distribution of each joint position $q_j$ in $\bb{q}$ is independent and follows a uniform distribution within its physical limits, $[q_{j_{min}}, q_{j_{max}}]$. We then estimate the distribution of end-effector positions by sampling joint positions in $\mathcal{Q}$. That is, we estimate $\mathcal{E}(\mathcal{C})$ by
\begin{align}
    \hat{\mathcal{E}}(\mathcal{C}) = \frac{1}{N} \sum\limits_{i = 1}^N E_{\theta}(f_e(\bb{q}_i \mid \mathcal{C})),
\end{align}
where $N$ is the number of samples, and the $\bb{q}_i$'s are uniformly sampled from $\mathcal{Q}$. Consequently, we reframe the MBPP into the following optimization problem
\begin{align}
    \textbf{[MBPP]:} \qquad & \arg \max \limits_{\mathcal{C}} \hat{\mathcal{E}}(\mathcal{C}), \\
    \text{s.t.} \qquad & \mathcal{C} \in \mathcal{B}.
\end{align}

We solve this problem by stochastic gradient ascent on $\hat{\mathcal{E}}$, where the gradient is computed as

\begin{align}
    {\partial \hat{\mathcal{E}} \over \partial \mathcal{C}} & = \frac{1}{N}  {\partial \over \partial \mathcal{C}} \sum\limits_{i=1}^M E_{\theta}(f_e(\bb{q}_i \mid \mathcal{C})) \\
    & = \frac{1}{N} \sum\limits_{i=1}^M {\partial E_{\theta}(f_e) \over \partial f_e} {\partial f_e(\bb{q}_i \mid \mathcal{C}) \over \partial \mathcal{C}}.
\end{align}

${\partial E_{\theta}(f_e) \over \partial f_e}$ can be easily computed by automatic differentiation, as $E_{\theta}(\cdot)$ is a neural network model. The derivative of the forward kinematics ${\partial f_e(\bb{q}_i | \mathcal{C}) \over \partial \mathcal{C}}$ is also easily attainable by the package PyTorch Kinematics. 

\begin{figure}[t]
    \centering
    \includegraphics[width=0.47\textwidth, trim=.05cm .05cm .05cm .05cm, clip]{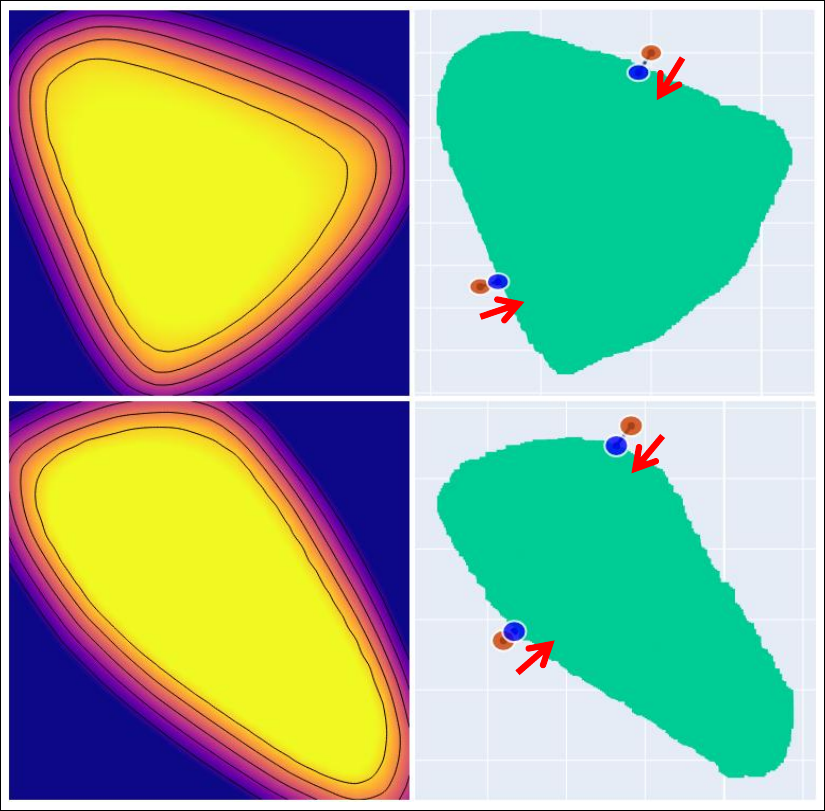}
    \caption{Left: Examples of constraint SIMs. Contours indicate equal output values. Right: Examples of applying \cref{alg:newtons-projection} to project infeasible points (brown) back into the feasible region. Red arrows show the direction of projection. The resulting points on the boundary are shown in blue.}
    \label{fig:projection}
\end{figure}

\begin{algorithm}[t]
    \caption{Projection to Boundary}
    \small
    \begin{algorithmic}[1]
    \small
        \State \textbf{Input}: constraint Spatial Instruction Map $\mathcal{G}(\cdot)$, coordinates $\bb{p} = (x, y)$, threshold probability $\tau$, maximum number of iterations $T_\text{Project}$, precision $\epsilon$
        \Function{Project}{}
            \For{$t \gets 1$ to $T_\text{Project}$}
                \If{$\lvert \mathcal{G}(\bb{p}) - \sigma^{-1}(\tau) \rvert > \epsilon$}
                    \State $\bb{g} \gets \nabla_\bb{p} \mathcal{G(\bb{p})}$
                    \State $\bb{p}\gets \bb{p} - {\bb{g} \over \lVert \bb{g} \rVert^2}(\mathcal{G}(\bb{p}) - \sigma^{-1}(\tau))$ \Comment{\cref{eq:project_back}}
                \Else
                    \State \textbf{break}
                \EndIf
            \EndFor
        \EndFunction
        \State \textbf{Output}: $\bb{p}$
    \end{algorithmic}
    \label{alg:newtons-projection}
\end{algorithm}

\subsection{Constraints} \label{subsec:constraints}
The discussion above derives an objective for the MBPP, where we wish the coverage of the manipulator over the regions of interest to be maximized. We now describe the imposition of the constraint $\mathcal{B}$ on the base pose. Specifically, constraints are separately applied to the 3D coordinates of the base $\bb{x} = (x, y, z)$ and its yaw angle $\omega$. We optimize with projected gradient methods \cite{book} to enforce constraint satisfaction, taking the following constraints into consideration:
\begin{enumerate}
\item \textbf{Diagrammatic instruction constraints on the $x,y$ coordinates of the base position:} If the human operator opts to specify a permissible region, $\mathcal{X}_p$, the same method discussed in \ref{subsec:region} is used to find the coordinates of the region in 3D, and their $x, y$ components are extracted. 

\item \textbf{Box constraint on $z$:} The height of a mobile base could be variable, e.g. on a quadruped, so we restrict $z \in [z_{min}, z_{max}]$.

\item \textbf{Box constraint on $\omega$:} Likewise, we restrict $\omega \in [\omega_{min}, \omega_{max}]$.
\end{enumerate}

During gradient ascent, if a step along the gradient results in an infeasible solution, the current solution is updated to be nearest feasible solution, i.e. the height $z$ is clamped to $[z_{min}, z_{max}]$, the yaw angle $z$ is clamped to $[\omega_{min}, \omega_{max}]$, and the $x, y$ coordinates are projected to the nearest feasible solution in $\mathcal{X}_p$, which can be efficiently handled by leveraging the method discussed in \ref{subsec:ebm}. Specifically, we train another constraint SIM $G(\cdot)$ to predict the probability $\bb{P}(\bb{p} \in \mathcal{X}_p \mid \bb{p})$, for any $\bb{p} \in \mathbb{R}^2$, by contrastive loss. The boundary of the region is then an equal-probability contour corresponding to a threshold $\tau$. If a gradient update steps out from $\mathcal{X}_p$, the $x, y$ coordinates are projected back to the contour corresponding a $P = \tau$. For added numerical stability, we take $\mathcal{G}(\cdot)$ as the output of $G(\cdot)$ prior to the final sigmoid layer, and instead solve for $\mathcal{G}(\bb{p}) = \sigma^{-1}(\tau)$, where $\sigma^{-1}(\cdot)$ is the inverse of sigmoid function. We use Newton's method towards the feasible region by considering the residual, 
\begin{equation}
    \bb{p} \leftarrow \bb{p} - {\nabla_\bb{p} \mathcal{G(\bb{p})} \over \lVert \nabla_\bb{p} \mathcal{G(\bb{p})} \rVert^2} (\mathcal{G}(\bb{p}) - \sigma^{-1}(\tau)). \label{eq:project_back}
\end{equation}
The projection of infeasible solutions to constraint satisfaction is described in Algorithm \ref{alg:newtons-projection}, and the entire optimization procedure is outlined in Algorithm \ref{alg:mbpp-solver}. Additionally, the SIMs and projection of a number of points outside the constraint area is shown in Fig. \ref{fig:projection}.

\begin{algorithm}[t]
    \caption{MBPP Optimizer}
    \small
    \begin{algorithmic}[1]
    \small
        \State \textbf{Input}: step size $\alpha$, initial base placement $\mathcal{C}_0 = (x_0, y_0, z_0, \omega_0)$, height limits $z_{min}, z_{max}$, yaw angle limits $\omega_{min}, \omega_{max}$, number of joint position samples $N$, allowable joint positions $\mathcal{Q}$, forward kinematics $f_e(\cdot)$, energy function $E_\theta(\cdot)$,  number of iterations $T_\text{MBPP}$, constraint Spatial Instruction Map $G(\cdot)$, threshold $\tau$.
        \State Initialize $\mathcal{C} \gets \mathcal{C}_0$
        \For{$t \gets 1$ to $T_\text{MBPP}$}
            \State Uniformly sample $N$ joint configs $\bb{q}_{1 \cdots N} \in \mathcal{Q}$
            \State $\hat{\mathcal{E}} \gets {1 \over N} \sum_{i=1}^N E_\theta(f_e(\bb{q}_i \mid \mathcal{C}))$
            \State $\bb{g} \gets \frac{1}{N} \sum_{i=1}^N {\partial E_{\theta}(f_e) \over \partial f_e} {\partial f_e(\bb{q}_i | \mathcal{C}) \over \partial \mathcal{C}}$
            \State $(x, y, z, \omega) \gets \mathcal{C} + \alpha \bb{g}$ \Comment{Gradient ascent on $\hat{\mathcal{E}}$}
            \State $z \gets clamp(z, z_{min}, z_{max})$
            \State $\omega \gets clamp(z, \omega_{min}, \omega_{max})$
            \State $\bb{p} \gets (x, y)$
            \If{$G(\bb{p}) < \tau$}
                \State $\bb{p} \gets$ \Call{Project}{$\bb{p}, \mathcal{G}, \tau$}
            \EndIf
            \State $(x, y) \gets \bb{p}$
            \State $\mathcal{C} \gets (x, y, z, \omega)$
        \EndFor
        \State \textbf{Output}: optimized base placement $\mathcal{C}$
    \end{algorithmic}
    \label{alg:mbpp-solver}
\end{algorithm}

\begin{figure}
    \centering
    \includegraphics[width=0.49\textwidth, trim=.5cm .5cm .5cm .5cm,clip]{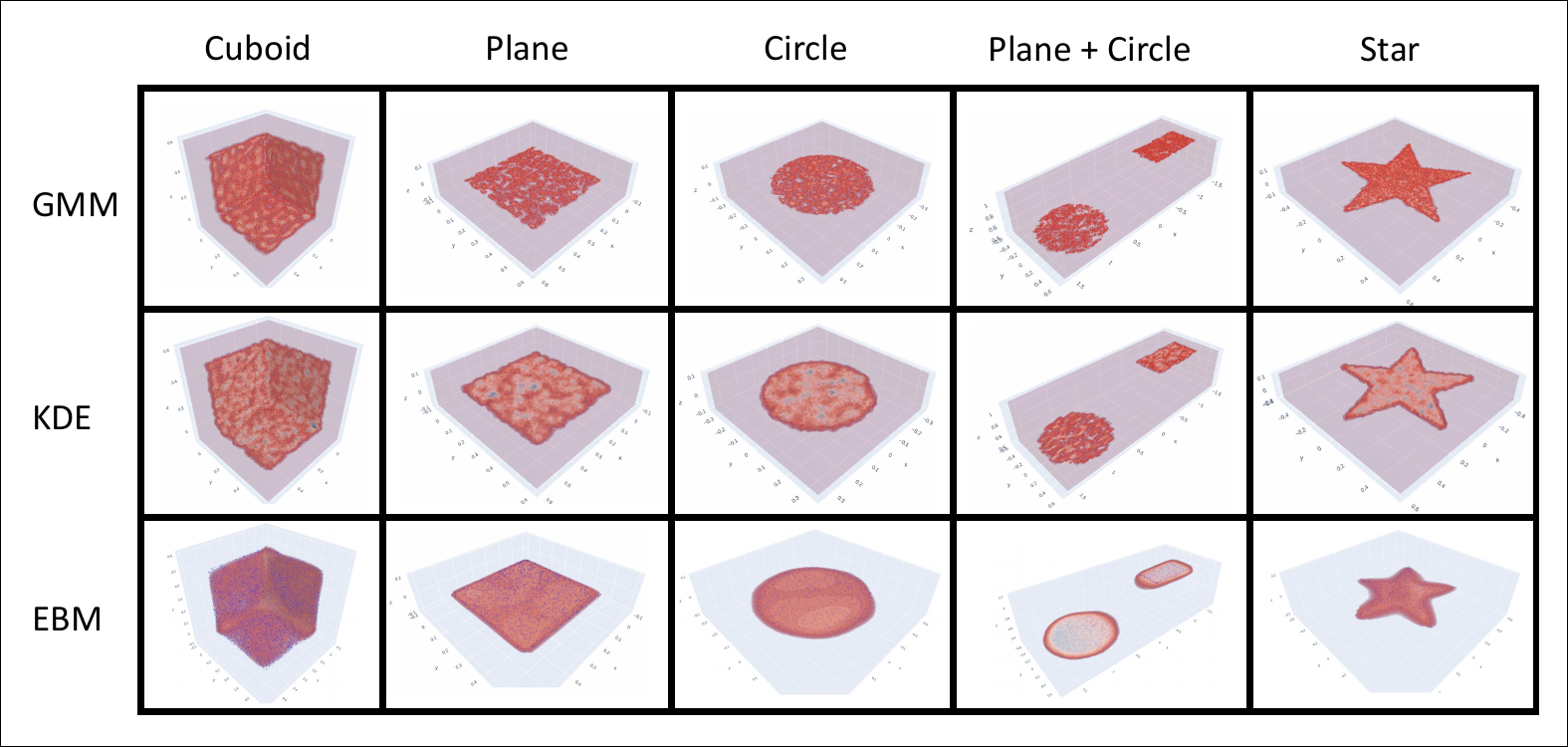}
    \caption{KDE, GMM, and our energy-based SIM fitted on different dataset of spatial points.}
    \label{fig:kde-gmm-ebm}
\end{figure}

\begin{figure*}[t]
    \centering
    \includegraphics[trim=.2cm .2cm .2cm .2cm, clip, width=0.95\textwidth]{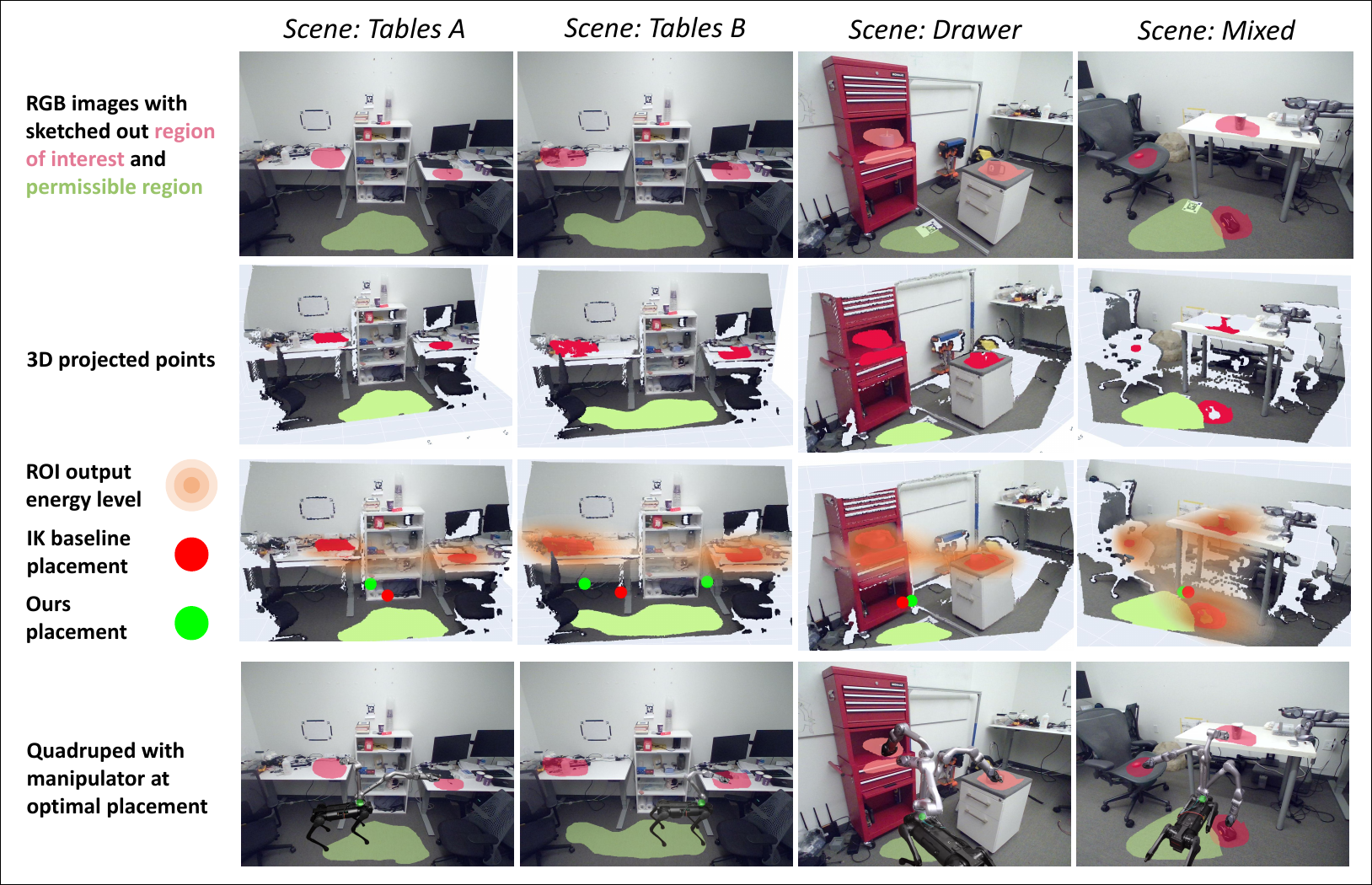}
    \caption{Row 1: Scenes with annotated SDIs. Row 2: SDIs projected into 3D. Row 3: Energy output levels from SIMs fitted on sketched ROIs. Row 4: Quadruped with manipulator posited at the optimal placements (indicated by green dot).}
    \label{fig:scenes}
\end{figure*}

\section{Empirical Evaluation}\label{sec: experiments}
In this section, we provide empirical evaluations on both the quality of the EBM representation of SIMs, as well as the quality of our optimization methodology. The specific questions that we seek to answer include: (1) How does the energy-based model (EBM) representation of Spatial Instruction Map (SIM) compare against alternative methods? (2) How does the optimization perform against baseline methods that solve for mobile base placement? (3) How does our approach handle multi-modality, or multiple specified regions which cannot all be covered?

\subsection{EBMs Represent SIMs Well}
We seek to investigate how our EBM representation of Spatial Diagrammatic Maps compares against other common approaches, in constructing continuous probability density estimates. We evaluate the quality of estimating a distribution over datapoints from the following datasets: \textbf{Cuboid}, \textbf{Plane}, \textbf{Circle}, \textbf{Plane + Circle} and \textbf{Star}, as shown in Fig. \ref{fig:kde-gmm-ebm}. For each dataset, we fit the model on the dataset, and report the per-sample log-likelihood on another dataset sampled from the same distribution.

In particular, we compare against the following baseline methods: \begin{itemize}
\item \textbf{Gaussian Mixture models (GMMs)} \cite{Bishop:2006}: We fit mixtures of Gaussians with full covariances. We evaluate on three variants of the GMM, each with 10, 50 and 100 Gaussians in the mixture.
\item \textbf{Kernel Density Estimator (KDE)} \cite{Silverman86}: We fit KDEs with a Gaussian kernel with a length-scale parameter tuned to maximize the log-likelihood.
\end{itemize}

Table \ref{tab:kde-gmm-ebm} summarizes the values of log likehood for KDE, GMM, and our SIMs that are built on EBMs. KDE yields worse results than GMM, while GMM-50 outperforms GMM-10 and also GMM-100 by a narrow margin. This suggests that increasing the number of Gaussians beyond a limit leads to overfitting and degrades the performance. On the other hand, the log likelihoods of EBMs are consistently above those of KDEs and GMMs, indicating that EBMs perform better at approximating those distributions and can generalize to spatial regions where observations are relatively sparse.
\begin{table}[t]
\centering
    \adjustbox{max width=0.49\textwidth}
    {
        \begin{tabular}{|c|ccccc|}
            \hline
             & KDE & GMM-10 & GMM-50 & GMM-100 & Ours \\
            \hline
            Cuboid & 3.851 & 6.034 & 6.182 & 6.177 & \bf{8.039} \\
            Plane & 4.909 & 7.213 & 7.217 & 7.119 & \bf 9.194 \\
            Circle & 5.169 & 7.490 & 7.506 & 7.389 & \bf 9.381 \\
            Plane + Circle & 2.859 & 5.596 & 5.612 & 5.465 & \bf 6.979 \\
            Star & 4.753 & 7.023 & 7.095 & 7.081 &  \bf 8.656 \\
            \hline
        \end{tabular}
    }
\caption{Our SIMs which leverage EBMs outperform KDE and GMM models, based on log-likelihood metrics. Higher log-likelihood values indicate better performance.}
\label{tab:kde-gmm-ebm}
\end{table}

\begin{table}[t]
    \centering
    \begin{tabular}{|c|c|c|}
        \hline 
        Hyperparameter & Value \\
        \hline 
        $N$ & $1024$ \\
        $T_\text{Project}$ & $20$ \\
        $T_\text{MBPP}$ & $40$ \\
        $\alpha$ & $0.005$ \\
        $z_{min}, z_{max}$ & $0.15 \, m, 0.42 \, m$ \\
        $\omega_{min}, \omega_{max}$ & $0, 2\pi$ \\
        $\mathcal{Q}, f_e(\cdot)$ & Specified by URDF \\
        $\tau$ & $0.95$ \\
        \hline
    \end{tabular}
    \caption{Hyperparameters used for the optimization.}
    \label{tab:hyperparameters}
\end{table}

\subsection{Solving the MBPP with Spatial Diagrammatic Maps}
We evaluate our approach of formulating and solving optimization problems which build upon our SIMs on multiple real-world scenes, where images are taken by a Vzense time-of-flight RGB-D camera. Diagrammatic instructions to specify ROIs for manipulation and permissible regions for the mobile base are sketched onto the images. This gives us 4 scenes, named \emph{Tables A}, \emph{Tables B}, \emph{Drawer}, and \emph{Mixed}.

We use the Unitree Z1 as the mobile manipulator the Unitree Aliengo quadruped as the mobile platform, which can elevate the manipulator base to a height between $0.15 \, m$ and $0.42 \, m$ above the ground. The quadruped is omnidirectional, i.e. $\omega \in [0, 2\pi)$. We use the joint limits specified in the URDF for Z1. We report the percentage of area of the ROI that is reachable via the simulator PyBullet \cite{coumans2019}. \cref{fig:scenes} shows the specified ROI and constraint overlayed on the images, the 3D projected points for each region, and the optimized base placements. The hyperparameters used for the optimization in our experiments are tabulated in \cref{tab:hyperparameters}.

We compare our method against a \textbf{Randomized Baseline} and an \textbf{Inverse Kinematics (IK) Baseline}. The randomized baseline randomly selects a base placement that satisfies all the constraints. Within the IK baseline, points are first sampled from the distribution learned by $E_\theta(\cdot)$, and for each point, Pybullet's IK solver is used to find, if any, a base placement and joint positions that make the end-effector reach that point. The output is the mean of all the base placements found, ensuring that the solution satisfies the constraints. \cref{tab:scores_v} reports the percentage of points reachable for all three methods, along with the runtimes.

\begin{table}[t]
    \centering
    \adjustbox{max width=0.49\textwidth}{
        \begin{tabular}{|c|c|c|c|c|}
            \hline
            Scene & \emph{Tables A} & \emph{Tables B} & \emph{Drawer} & \emph{Mixed} \\
            \hline
            Random reachability (\%) & 1.123 & 2.588 & 59.668 & 7.178 \\
            IK reachability (\%) & 1.660 & 2.832 & 70.263 & 15.039 \\
            Ours reachability (\%) & \bf 30.469 & \bf 50.000 & \bf 77.637 & \bf 61.865 \\
            \hline
            IK runtime (s) & 2.234 & 2.417 & 2.306 & 2.113 \\
            Ours runtime (s) & \bf 0.350 & \bf 0.424 & \bf 0.440 & \bf 0.451 \\
            \hline
        \end{tabular}
    }
    
    \caption{The performance of random baseline, IK baselines, and ours, evaluated through percent reachability and runtime. Higher reachability indicates better performance while lower runtime indicates greater efficiency.}\label{tab:scores_v}
\end{table}
We observed that the IK baseline suffers from long runtime and suboptimal results. This is particularly the case in \emph{Tables B}, where there exists multiple ROIs that are far apart. Additionally, the IK solution may often not produce significantly better results than the random baseline. This is because the inverse kinematics computes the base placement solely by a binary state indicating whether a position is reachable. That is, it does not incorporate the \emph{probability} that a position is a region of interest.

\subsection{Multi-modality in Diagrammatic Instructions}
\label{subsec:multimodality}
A mobile manipulator may have distinct ROIs that are separated by considerable distance. The SIM and the solution to the MBPP should be able to handle multiple distinct modes over the space of the environment. The solution to the environment \emph{Tables B}, shown in \Cref{fig:scenes}, demonstrates the multi-modality. We observe that our approach to the MBPP can produce distinct solutions that provide coverage for each specific mode, while the inverse kinematics baseline is conflicted and tries unsuccessfully to cover both regions, resulting in a lower coverage percentage. On the other hand, when two ROIs can be simultaneously covered by a single placement of the manipulator, such as in \emph{Tables A}, our approach finds the base position that best covers both regions.

\section{Conclusions and Future Work} \label{sec:conc}
In this work, we build a \emph{Spatial Diagrammatic Instructions} framework, where the human operator provides Spatial Diagrammatic Instructions to convey to the robot regions that are relevant for an instruction. We propose \emph{Spatial Instruction Maps}, which learn differentiable and probabilistic models that represent the relevant 3D regions. We demonstrate that our spatial representations faithfully capture the spatial regions. We also show that their differentiability empowers efficient optimization for downstream tasks by applying the representation on the problem of optimal placement of a mobile manipulator, and the result of optimization achieves a high coverage over the regions specified by the user. Therefore, we envision an avenue of future research to be adapting our Spatial Instruction Maps to be not only spatial but also \emph{spatiotemporal}, enabling time dependent instructions to be incorporated.

{\small
\bibliographystyle{IEEEtran}
\bibliography{refs}
}

\end{document}